%% file: PaperForReview.tex
\documentclass[10pt,twocolumn,letterpaper]{article}

\usepackage{cvpr}              % To produce the CAMERA-READY version

\usepackage{float}
\usepackage{graphicx}
\usepackage{amsmath,mathrsfs}
\usepackage{amssymb}
\usepackage{booktabs}
\usepackage{multirow}
\usepackage{bbding}
\usepackage[accsupp]{axessibility}
\usepackage[table,xcdraw]{xcolor}
\usepackage[linesnumbered,ruled,vlined,lined,boxed,commentsnumbered]{algorithm2e}

\definecolor{cvprblue}{rgb}{0.21,0.49,0.74}
\usepackage[pagebackref,breaklinks,colorlinks,citecolor=cvprblue]{hyperref}

\usepackage{adjustbox}

\usepackage{xcolor} % For the \textcolor command
\usepackage{soul} % For the \hl command

\usepackage[pagebackref,breaklinks,colorlinks]{hyperref}

\usepackage[capitalize]{cleveref}
\crefname{section}{Sec.}{Secs.}
\Crefname{section}{Section}{Sections}
\Crefname{table}{Table}{Tables}
\crefname{table}{Tab.}{Tabs.}

\definecolor{generalized}{HTML}{C6FEDD}
\definecolor{finetuning}{HTML}{FBD785}
\definecolor{instance}{HTML}{F64F58}

\def\paperID{13636} % *** Enter the Paper ID here
\def\confName{CVPR}
\def\confYear{2024}

\begin{document}

\vspace{-0.4in}
\title{GP-NeRF: Generalized Perception NeRF for Context-Aware \\ 3D Scene Understanding}

\author{
    Hao Li$^{1}$, Dingwen Zhang$^{1,6,*}$, Yalun Dai$^4$, Nian Liu$^{2,*}$, Lechao Cheng$^3$ ,  Jingfeng Li$^1$, \\ Jingdong Wang$^5$, Junwei Han$^{1,6}$\\
    $^1$ Brain and Artificial Intelligence Lab, Northwestern Polytechnical University
    $^2$ MBZUAI \\
    $^3$ Hefei University of Technology
    $^4$ Nanyang Technological University $^5$  Baidu, Inc.\\
    $^6$ Institute of Artificial Intelligence, Hefei Comprehensive National Science Center \\
    $^*$  Corresponding authors
}

 \twocolumn[{%
 \renewcommand\twocolumn[1][]{#1}%
 \maketitle
 \begin{center}
 \vspace{-0.4in}
 \centering\includegraphics[width=0.9\textwidth]{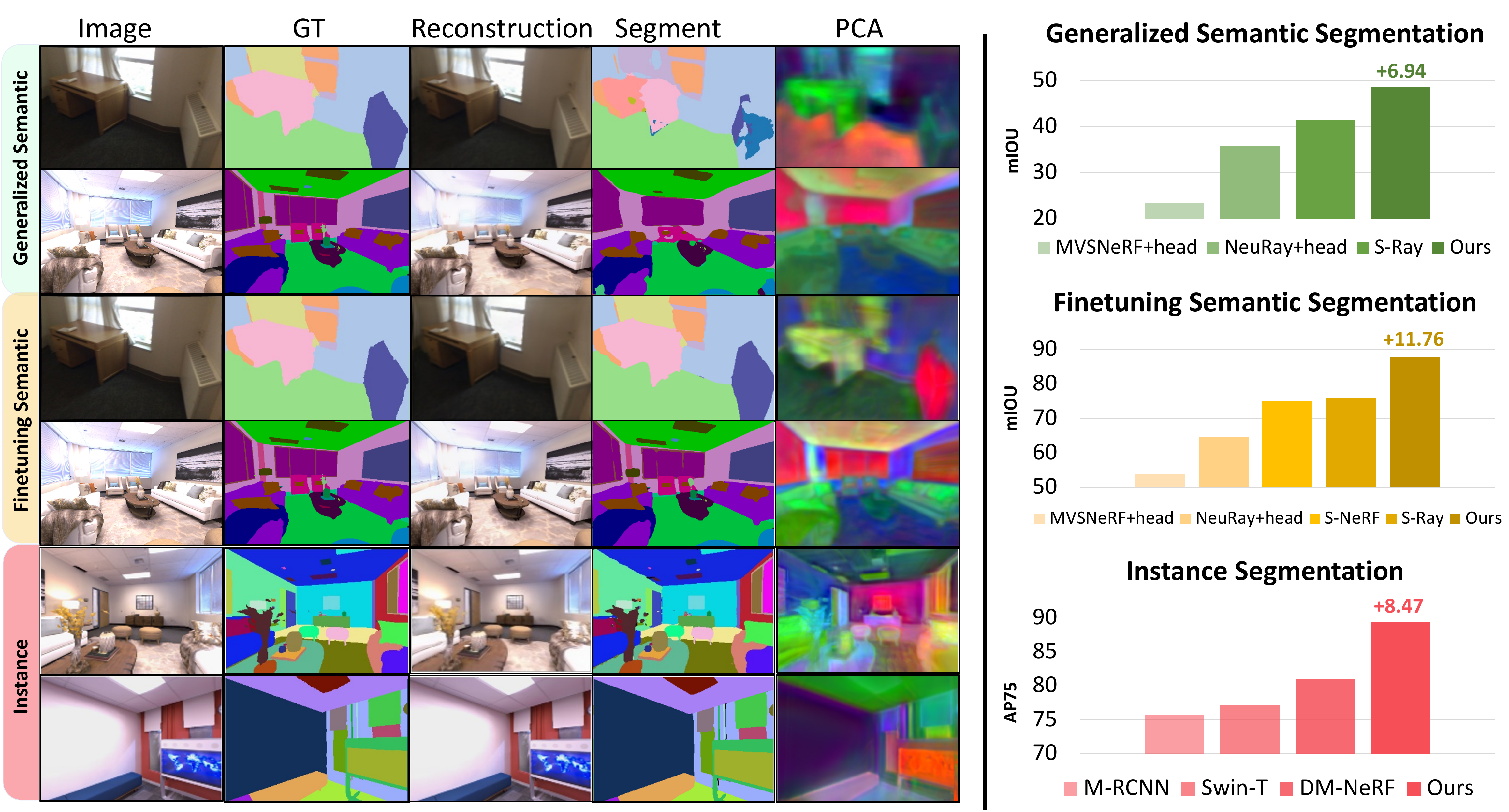}
 \captionof{figure}{Our method, called \textbf{GP-NeRF}, achieves remarkable performance improvements for instance and semantic segmentation in both synthesis~\cite{straub2019replica} and real-world~\cite{dai2017scannet} datasets, as shown in the right column of the figure. 
 Here we showcase \sethlcolor{generalized}\hl{generalized semantic segmentation}, \sethlcolor{finetuning}\hl{finetuning semantic segmentation}, and \sethlcolor{instance}\hl{instance segmentation}) with their corresponding reconstruction results. 
 For the left column, the qualitative results of the visualization are presented, showing the effectiveness of our method for simultaneous segmentation and reconstruction. 
 What's more, we visualize our rendered features via PCA in the novel view, demonstrating our method possesses the capability to produce semantic-aware features that can distinguish between different classes and objects. 
 % For the right column, the qualitative results show that our method achieves the SOTA performance on all three segmentation settings.
 }
 \end{center}%
 }]

%%%%%%%%% ABSTRACT
\begin{abstract}
Applying Neural Radiance Fields (NeRF) to downstream perception tasks for scene understanding and representation is becoming increasingly popular. 
Most existing methods treat semantic prediction as an additional rendering task, \textit{i.e.}, the "label rendering" task, to build semantic NeRFs.
However, by rendering semantic/instance labels per pixel without considering the contextual information of the rendered image, these methods usually suffer from unclear boundary segmentation and abnormal segmentation of pixels within an object. 
To solve this problem, we propose Generalized Perception NeRF (GP-NeRF), a novel pipeline that makes the widely used segmentation model and NeRF work compatibly under a unified framework, for facilitating context-aware 3D scene perception. To accomplish this goal, we introduce transformers to aggregate radiance as well as semantic embedding fields jointly for novel views and facilitate the joint volumetric rendering of both fields.
In addition, we propose two self-distillation mechanisms, i.e., the Semantic Distill Loss and the Depth-Guided Semantic Distill Loss, to enhance the discrimination and quality of the semantic field and the maintenance of geometric consistency.
In evaluation, as shown in Fig. 1 we conduct experimental comparisons under two perception tasks (\textit{i.e.} semantic and instance segmentation) using both synthetic and real-world datasets. Notably, our method outperforms SOTA approaches by 6.94\%, 11.76\%, and 8.47\% on generalized semantic segmentation, finetuning semantic segmentation, and instance segmentation, respectively. \href{https://lifuguan.github.io/gpnerf-pages/}{Project}.
%The experimental results demonstrate that our proposed method can significantly surpass current perception NeRF methods in terms of both scene perception and reconstruction accuracy.
\end{abstract}

%%%%%%%%% BODY TEXT
\section{Introduction}
\label{sec:intro}

Robust scene understanding models are crucial for enabling various applications, including virtual reality (VR)~\cite{jaritz2019multi}, robot navigation~\cite{ye2023pvo}, self-driving~\cite{feng2020deep}, and more~\cite{awais2023foundational}. They have experienced tremendous progress over the past years, driven by continuously improved model architectures~\cite{mask2former,maskformer,knet, liu2023multi,liu2022intermediate} in 2D image segmentation. However, these methods face challenges due to their lack of specific scene representation and the inability to track unique object identities across different views~\cite{siddiqui2023panopticlifting}.

Meanwhile, implicit neural representations~\cite{mildenhall2021nerf,liu2022neuray,t2023GNT,wang2021ibrnet} have demonstrated an impressive capability in capturing the 3D structure of complex real-world scenes~\cite{dai2017scannet}. By adopting multi-layer perceptions, it utilizes multi-view images to learn 3D representations for synthesizing images in novel views with fine-grained details. 
This success has spurred research into applying NeRF for robust scene understanding, aiming to explore a broader range of possibilities in high-level vision tasks and applications.

Recent works~\cite{zhi2021semanticnerf,liu2023semanticray,fu2022panopticnerf,siddiqui2023panopticlifting} addressed scene understanding from 2D images by exploring semantics using Neural Radiance Fields (NeRFs)~\cite{mildenhall2021nerf}.
%nian: semantic NerF的出发点是用nerf做语义分割吗？说反了吧，不因该是在nerf里面做语义分割吗？
Per-scene optimized methods, such as Semantic-NeRF~\cite{zhi2021semanticnerf}, DM-NeRF~\cite{wang2022dm}, and Panoptic-NeRF~\cite{fu2022panopticnerf},  simply utilize additional Multi-Layer Perceptron (MLP) to regress the semantic class for each 3D-point together with radiance and density. The latest method Semantic-Ray~\cite{liu2023semanticray}, based on generalized NeRF NeuRay~\cite{liu2022neuray}, achieves generalized semantic segmentation by introducing an individual learnable semantic branch to construct the semantic field and render semantic features in novel view using frozen density.

Although this operation is reasonable to build a semantic field, it falls short in achieving joint optimization of both RGB rendering and semantic prediction, thus missing an important message when building high-quality heterogeneous embedding fields: 
The geometry distribution of the radiance field and Semantic-Embedding field should be consistent with each other.
For example:
 1) The boundaries of different objects are usually distinct in RGB representation, they could be utilized for achieving more accurate boundary segmentation; 
 and 2) The areas belonging to the same object often share consistent coloration, which can act as informative cues to enhance the quality of RGB reconstruction. Moreover, Semantic-Ray follows the vanilla semantic NeRF by rendering semantic labels for each point independently in the novel view, ignoring the context information, such as the relationships and interactions between the nearby pixels and objects.

\begin{figure*}[htbp]
  \centering
  \includegraphics[width=0.95\linewidth]{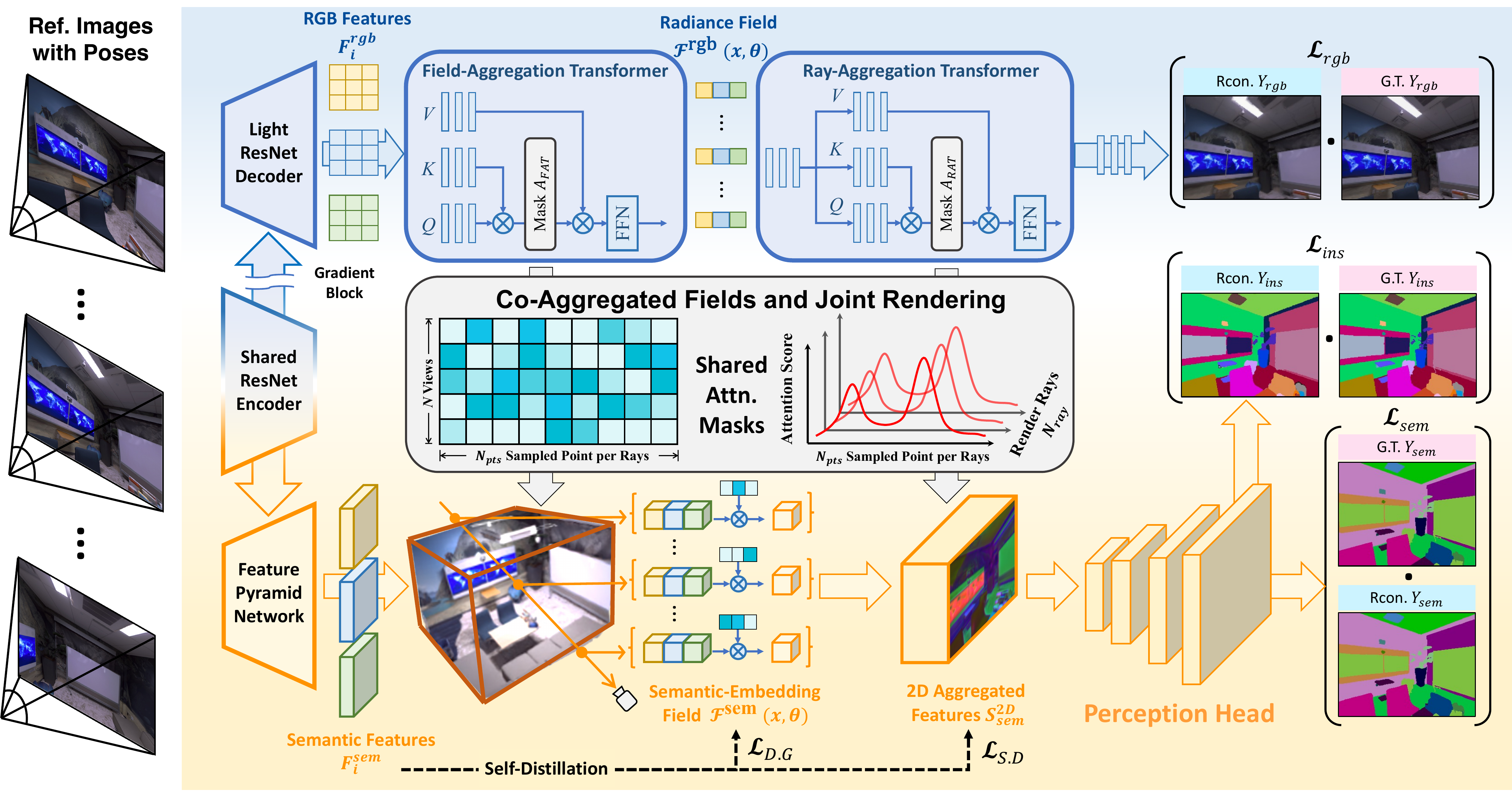}
  \vspace{-0.05in}
  \caption{\textbf{Overview of proposed GP-NeRF}. Given reference views with their poses, we embed NeRF into the segmenter to perform context-aware semantic $Y_{sem}$ /instance $Y_{ins}$ segmentation and ray reconstruction $Y_{rgb}$ in novel view (Sec. \ref{sec:framework}).  In detail, we use Transformers to co-aggregate {\color[HTML]{004C99} Radiance}  as well as {\color[HTML]{FF9C38} Semantic-Embedding} fields and render them jointly in novel views (Sec. \ref{subsec:fields}). Specifically, we propose two self-distillation mechanisms to boost the discrimination and quality of the semantic embedding field (Sec. \ref{subsec:optimization}). }
  \label{fig:overview}
    \vspace{-0.25in}
\end{figure*}

To address these problems, we present Generalized Perception NeRF (GP-NeRF), a novel unified learning framework that embeds NeRF and the powerful 2D segmentation modules together to perform context-aware 3D scene perception. 
As shown in Fig. \ref{fig:overview}, GP-NeRF utilizes Field Aggregation Transformer to aggregate the radiance field as well as the semantic-embedding field, and Ray Aggregation Transformer to render them jointly in novel views. Both processes perform under a joint optimization scheme.
Specifically, we render rich-semantic features rather than labels in novel views and feed them into a powerful 2D segmentation module to perform context-aware semantic perception.
% As shown in Fig. \ref{fig:overview}, GP-NeRF contains a Transformer-based Ray Rendering Branch (RRB) and a Context-aware Semantic Rendering Branch (SRB), where the two branches perform under a joint optimization scheme. Specifically, the RRB is designed to construct scene representations and render novel views on a per-pixel manner, while the SRB is used to build semantic embedding field upon the novel-view scene representations and the rich context features to perform the semantic prediction. 
To enable our framework to work compatibly, we further introduce two novel self-distillation mechanisms: 1) the Semantic Distill Loss, which enhances the discrimination and quality of the semantic field, thereby facilitating improved prediction performance by the perception head; and 2) the Depth-Guided Semantic Distill Loss, which aims to supervise the semantic representation of each point within the semantic field, ensuring the maintenance of geometric consistency. Under such mechanisms, our method bridges the gap between the powerful 2D segmentation modules and NeRF methods, offering a possible integration solution with existing downstream perception heads.

%We conduct detailed experiments on synthesis~\cite{straub2019replica} and real-world~\cite{dai2017scannet} datasets to evaluate our performance in both generalized and per-scene optimization settings. As evidenced by our experiments, this pipeline is able to represent the semantic embedding across scenes and jointly optimize both ray rendering and scene perception. It surpasses existing NeRF methods in both reconstruction quality and perception accuracy on both instance and semantic tasks.  

Our contributions can be summarized as follows:
\begin{itemize}
    \item We make an early effort to establish a unified learning framework that can combine NeRF and segmentation modules to perform context-aware 3D scene perception.
    \item Technically, we use Transformers to jointly construct radiance as well as semantic embedding fields and facilitate the joint volumetric rendering upon both fields for novel views.
    \item The 2D and depth-guided self-distillation mechanisms are proposed to boost the discrimination and quality of the semantic embedding field.
    \item Comprehensive experiments are conducted. The results demonstrate that our method can surpass existing NeRF methods in downstream perception tasks (\textit{i.e.}semantic, instance) with both generalized and per-scene settings.
\end{itemize}

\section{Related Work}

\subsection{Neural Radiance Fields (NeRF)}
% Neural Radiance Fields (NeRF) introduced by ~\cite{mildenhall2021nerf} synthesize consistent and photorealistic novel views by fitting each scene as a continuous 5D radiance field parameterized by an MLP. Since then, several works have improved NeRFs further. For example, Mip-NeRF ~\cite{barron2021mip, barron2022mip} efficiently addresses scale of objects in unbounded scenes, Nex ~\cite{wizadwongsa2021nex} models large view-dependent effects, others ~\cite{oechsle2021unisurf, yariv2021volume} improve the surface representation, extend to dynamic scenes ~\cite{park2021nerfies, pumarola2021d}, introduce lighting and reflection modeling ~\cite{chen2021nerv, verbin2022ref}, or leverage depth to regress from few views ~\cite{xu2022sinnerf, deng2022depth}. Other works aim to avoid per-scene training, similar to PixelNeRF ~\cite{yu2021pixelnerf}, IBRNet ~\cite{wang2021ibrnet}, NeuRay ~\cite{liu2022neuray}, and GNT~\cite{t2023GNT}, which train a cross-scene multi-view aggregator and reconstruct the radiance field with a one-shot forward pass. 
% Based on cross-scene Nerf, we designed a Generalized Semantic and Render Joint Field to simultaneously complete cross-scene reconstruction and segmentation tasks.

Neural Radiance Fields (NeRF), introduced by Mildenhall et al. ~\cite{mildenhall2021nerf}, have revolutionized view synthesis by fitting scenes into a continuous 5D radiance field using MLPs. Subsequent enhancements include Mip-NeRF's ~\cite{barron2021mip, barron2022mip} efficient scaling in unbounded scenes, Nex's ~\cite{wizadwongsa2021nex} handling of large view-dependent effects, improvements in surface representation~\cite{oechsle2021unisurf, yariv2021volume} and dynamic scene adaptation ~\cite{park2021nerfies, pumarola2021d}, as well as advancements in lighting, reflection ~\cite{chen2021nerv, verbin2022ref}, and depth-based regression ~\cite{xu2022sinnerf, deng2022depth}. Methods like PixelNeRF ~\cite{yu2021pixelnerf}, IBRNet ~\cite{wang2021ibrnet}, NeuRay ~\cite{liu2022neuray}, and GNT ~\cite{t2023GNT} further reduce the need for per-scene training by using cross-scene multi-view aggregators for one-shot radiance field reconstruction. 
Building on these cross-scene Nerf methods, our work introduces a generalized semantic and rendering joint field, aiming to achieve simultaneous cross-scene reconstruction and segmentation.

\subsection{NeRFs with Scene Understanding}
Encoding semantics into NeRF is essential for scene understanding. 
Semantic NeRF~\cite{zhi2021semanticnerf} first explored introducing vanilla NeRF into semantic masks by adopting extra MLP layers to "render" semantic labels.
DFF~\cite{dff} and FeatureNeRF~\cite{ye2023featurenerf} utilize the pre-trained CLIP network and employ extra MLP layers for distillation learning to render text-aligned semantics features. 
Panoptic-Lifting~\cite{siddiqui2023panopticlifting} directly distills labels from Mask2former's predicted probabilities~\cite{mask2former}.
Based on Generalize NeRF~\cite{liu2022neuray}, Semantic-Ray~\cite{liu2023semanticray} adds an additional semantic branch to perform per-pixel semantic label rendering.

% More recently, NeRF-RPN~\cite{nerfrpn} uses pre-trained NeRF to present multi-scale 3D neural volumetric features and adopts 3D RPN to detect 3D objects in a scene. It gains competitive performance while limiting the reconstructed performance because the NeRF model is frozen.

In conclusion, although these methods have extended the idea, \textit{e.g.}, by applying to panoptic tasks~\cite{fu2022panopticnerf}, adding large language model (LLM)~\cite{clip} features~\cite{dff,cen2023segment,ye2023featurenerf}, and making it generalize~\cite{liu2023semanticray}, they all consider the semantic problem as another "rendering" variant: {they render labels or features for each pixel independently, ignoring the contextual consistency among pixels in the novel view}.

In contrast to previous approaches, we frame the segmentation issue as {“prediction with context”} rather than “isolated label rendering”. Accordingly, we generate semantic-aware features instead of labels from our semantic-embedding field in new views. Moreover, we are able to perform context-aware segmentation thanks to the capabilities of the segmenter, which is a feature that previous methods lacked.  Thanks to this design, the rendering and segmentation branches can benefit each other. 
Therefore, unlike ~\cite{nerfrpn}, which enhances 3D object detection performance at the expense of reconstruction performance, our method can simultaneously improve both reconstruction and segmentation performance.
\vspace{-0.05in}
\section{Preliminaries}
\label{sec:Preliminaries}
In this section, we take a brief review of  GNT~\cite{t2023GNT}.
NeRF represents a 3D scene as a radiance field \(\mathcal{F}:(\boldsymbol{x}, \boldsymbol{\theta}) \mapsto (\textbf{c}, \sigma)\), which maps the spatial coordinate $\textbf{x}$ to a density $\sigma$ and color $\textbf{c}$. While GNT models 3D scene as a coordinate-aligned feature field \(\mathcal{F}:(\boldsymbol{x}, \boldsymbol{\theta}) \mapsto \boldsymbol{f} \in \mathbb{R}^d\), $d$ is the dimension of the features. To learn this representation, GNT uses Transformer as a set aggregated function $\mathcal{V}(\cdot)$ to aggregate the features of reference views into a coordinate-aligned feature ﬁeld, which is formulated below:
\begin{equation}
    \mathcal{F}(\boldsymbol{x}, \boldsymbol{\theta})=\mathcal{V}\left(\boldsymbol{x}, \boldsymbol{\theta} ;\left\{\boldsymbol{I}_1, \cdots, \boldsymbol{I}_N\right\}\right)
\end{equation}

Subsequently, to obtain the final outputs $\boldsymbol{C}$ of the ray $\boldsymbol{r} = (\boldsymbol{o},\boldsymbol{d})$ in target view in this feature field, GNT parameterizes the ray by $\boldsymbol{r}(t) = \boldsymbol{o} +t\boldsymbol{d}$, $t\in[t_1,t_M]$, and uniformly samples  $M$ points $\boldsymbol{x}_i$ of feature representations $\boldsymbol{f}_i=\mathcal{F}\left(\boldsymbol{x}_i, \boldsymbol{\theta}\right) \in \mathbb{R}^d$ along the ray $\boldsymbol{r}$. Then, 
GNT adopts Transformer as a formulation of weighted aggregation to achieve volume rendering:
\begin{equation}
 \vspace{-0.05in}
    \boldsymbol{C}(\boldsymbol{r})= \text{MLP}  \circ  
\frac{1}{M} \sum^{M}_{i=1} A(x_{i})f(x_i, \theta),
 \vspace{-0.05in}
\end{equation}
where $A(x_i)$ is the weight of point $x_i$ computed by Transformer and $\boldsymbol{C}(\boldsymbol{r})$ is the rendered color of the ray $\boldsymbol{r}$.

\section{Methodology}
\label{sec:method}
%This section can be divided into three parts:
%(1) We introduce our \textbf{Overall Framework}(Sec. \ref{sec:framework}). Distinguishing from previous methods that directly render semantic labels, we embed NeRF into segmenters and perform photorealistic synthesis and context-aware perception in novel views.
%(2) We introduce our \textbf{Joint Field Representation and Rendering}(Sec. \ref{subsec:fields}) for ray and semantic feature rendering. Differing from previous approaches that separately construct radiance and semantic fields, we use Transformer modules to jointly construct radiance and semantic-embedding fields and render them in the novel views. 
%(3) Correspondingly, we leverage self-distillation mechanism \textbf{Semantic Distill Loss} and \textbf{Depth Guided Distill Loss}(Sec. \ref{subsec:optimization}) to boost the discrimination of rendered features as well as semantic embedding fields and perform semantic consistency in 3D grid to benefits our geometry awareness.

\subsection{Overall Framework}
\label{sec:framework}
Given \(N\) images $I=\left\{I_i \in\right.\left.\mathbb{R}^{H \times W \times 3}\right\}$ with corresponding poses, the training targets are to conduct scene perception (semantic $Y_{sem}$, instance $Y_{ins}$)  and reconstruction $Y_{rgb}$ in the novel \textit{target} views, where $Y_{sem}=\left\{Y_i \in\right.\left.\mathbb{R}^{H \times W \times O}\right\}$, $Y_{ins}=\left\{Y_i \in\right.\left.\mathbb{R}^{H \times W \times C}\right\}$, and $Y_{rgb}=\left\{Y_i \in\right.\left.\mathbb{R}^{H \times W \times 3}\right\}$, where \(O\) and \(C\) denote the number of semantic classes and instances. Unlike previous Semantic NeRF methods that directly render colors and semantic labels in a per-pixel manner, we perform segmentation tasks using (implicit) image context (Fig. \ref{fig:overview}). To accomplish this objective, we utilize NeRF to aggregate novel view semantic features \(\boldsymbol{S}^{2D}_{sem}\) from reference features \(\boldsymbol{F}^{sem}_i\)(Sec. \ref{subsec:fields}), where \(\boldsymbol{F}^{sem}_i\) is extracted by Multi-Scale Feature Extractor. After that, semantic features \(\boldsymbol{S}^{2D}_{sem}\) are fed into the Context-Aware Perception Head to perform image-wise context-aware perception.

\noindent \textbf{Multi-Scale Feature Extractor.} 
% Semantic-Ray uses single 3D contextual space to construct not only semantic but also radiance rendering. This approach constrains the semantic contextual awareness, as radiance rendering tends to prioritize low-level over high-level semantic features.
% Benefiting from our coordinate-aligned feature field that aggregate radiance field and semantic-embedding field seperately, we are enable to 
To enhance the semantic-aware of our semantic-embedding fields, for each reference image \(\boldsymbol{I}_i\), we use shared ResNet-34 followed by a Feature Pyramid Network (FPN)~\cite{lin2017fpn} module to produce multi-scale features \(\boldsymbol{F}^{sem}_i\) for our semantic field aggregation.

\noindent \textbf{Context-Aware Perception Head.} Our perception head takes rendered semantic features \(\boldsymbol{S}^{2D}_{sem}\) and outputs semantic labels \(Y_{sem}\) in novel view. 
Here we split \(\boldsymbol{S}^{2D}_{sem}\) into 4 parts \([s^{2D}_{sem,1}, s^{2D}_{sem,2}, s^{2D}_{sem,3}, s^{2D}_{sem,4}]\) to decompose high-level features and low-level features,
% With the aforementioned rich semantic features, we can easily integrate any state-of-the-art semantic network to boost our segmentation performance, whether it's based on Transformer~\cite{maskformer,mask2former,xie2021segformer} or CNN~\cite{unet,deeplabv3}. 
and adopt the decoder of the U-Net~\cite{unet} to verify our architecture's performance. In specific, for \textit{i}-th layer, it consists of an upsampling (\textit{i}-1)-th layer's output feature \(s'_{i-1}\) with \(2\times 2\) convolution("up-convolution"), a concatenation of \textit{i}-th feature map \(s^{2D}_{sem,i}\), and two \(3\times 3\) convolutions followed by a ReLU. The process can be formulated as below:
\begin{equation}
    s'_i = \text{ReLU} \cdot \text{Conv}(s^{2D}_{sem,i} + \text{Up-Conv}(s'_{i-1}))
\end{equation}

\begin{figure}[t]
  \centering
  \includegraphics[width=1\linewidth]{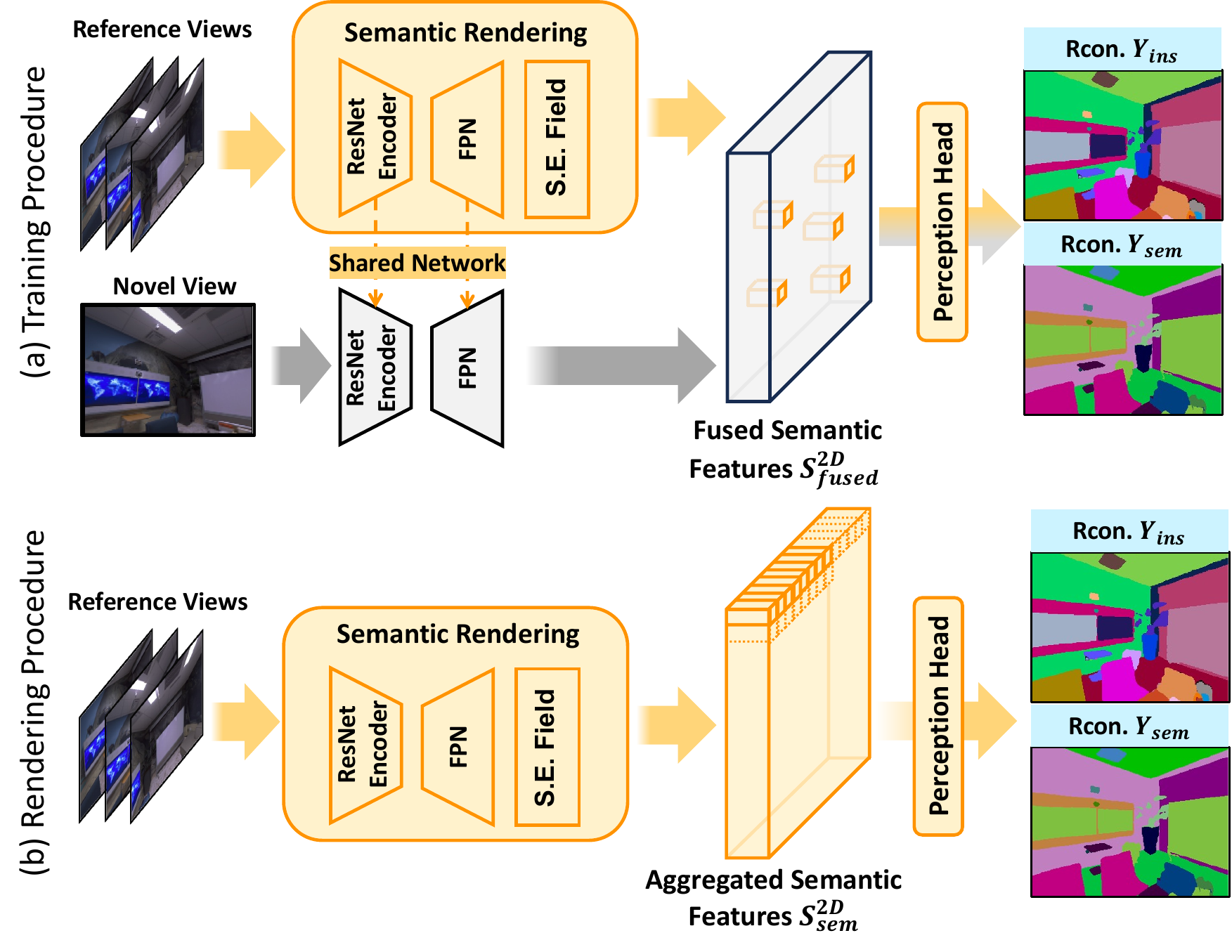}
 \vspace{-0.2in}
  \caption{Illustration of \textbf{training}(a) and \textbf{rendering}(b) procedure, where S.E. field denotes Semantic-Embedding Field.}
  \label{fig:implementation}
 \vspace{-0.2in}
\end{figure}

\noindent \textbf{Rendering and Training Process.} NeRF can only render limited \(N_{pts}\) points in each iteration, the same as our method.
During rendering, we stack all the semantic features \(\boldsymbol{S}^{2D}_{sem}(r)\) of sampled points as image-level features and feed them into the perception head together (see Fig. \ref{fig:implementation}(b)).
However, it's impossible to use fully rendered semantic features in every training batch. Therefore, as shown in Fig. \ref{fig:implementation}(a), for the semantic 2D map \(\boldsymbol{S}^{2D}_{sem}\), we specifically fill its unrendered areas with the corresponding regions from the novel 2D map \(\boldsymbol{S}^{2D}_{novel}\). This process creates a fused image-level feature map \(\boldsymbol{S}^{2D}_{fused}\), which is subsequently fed into the Perception Head for semantic prediction.
 \vspace{-0.1in}
\subsection{Co-Aggregated Fields and Joint Rendering}
\label{subsec:fields}

Given low-level features $\boldsymbol{F}^{rgb}_i$ and high-level features $\boldsymbol{F}^{sem}_i$ from Multi-Scale Feature Extractor, we use shared-attention(\textit{i.e.} Field-Aggregation Transformer) to co-aggregate the radiance field and semantic-embedding field. Subsequently, another shared-attention (\textit{i.e.} Ray-Aggregation Transformer) employs joint volumetric rendering from both fields to generate point-wise colors and semantic features in the novel view.
% , noticed that we aggregate semantic features rather than labels in .

\noindent \textbf{Co-Aggregate Radiance and Semantic-Embedding Fields.}  We represent a 3D scene as a coordinate-aligned feature field~\cite{t2023GNT}, which can attach low-level features for ray rendering or high-level features for scene understanding. 
Therefore, to obtain feature representations of position $\boldsymbol{x}$ in novel view,  following the idea of epipolar geometry constraint~\cite{suhail2022light}, $\boldsymbol{x}$ is projected to every reference image and interpolated the feature vector on the image plane. 
Firstly, the Field Aggregation Transformer (dubbed $\text{FAT}(\cdot)$) is adopted to combine all features $\boldsymbol{F}^{rgb}_i$ from reference views for radiance field $\mathcal{F}^{rgb}(\boldsymbol{x}, \boldsymbol{\theta})$ aggregation. Formally, this process can be written as:
\begin{equation}
\begin{aligned}
\mathcal{F}^{rgb}(\boldsymbol{x}, \boldsymbol{\theta}), \mathcal{A}_{FAT} = \text {FAT}( & \boldsymbol{F}^{rgb}_1\left(\Pi_1(\boldsymbol{x}), \boldsymbol{\theta}\right), \cdots, \\
& \boldsymbol{F}^{rgb}_N\left(\Pi_N(\boldsymbol{x}), \boldsymbol{\theta}\right)),
\end{aligned}
\label{eq:view}
\end{equation}
where  \(\Pi_i(\boldsymbol{x})\) projects $\boldsymbol{x}$ to \textit{i}-th reference image plane by applying extrinsic matrix, \(\boldsymbol{F}^{rgb}_i(\Pi_i(\boldsymbol{x}), \boldsymbol{\theta}) \in \mathbb{R}^{D_{rgb}}\) computes the feature vector at projected position \(\Pi_i(\boldsymbol{x}) \in \mathbb{R}^2\) via bilinear interpolation on the feature grids. Furthermore, $\mathcal{A}_{FAT}\in \mathbb{R}^{N_{pts}\times N}$ is the aggregation weight from Field Aggregation Transformer, which enables us to construct semantic embedding field $\mathcal{F}^{sem}(\boldsymbol{x}, \boldsymbol{\theta})$ easily by applying dot-product with features $\boldsymbol{F}^{sem}_i$ from reference views:
\begin{equation}
\begin{aligned}
  \mathcal{F}^{sem}(\boldsymbol{x}, \boldsymbol{\theta}) = & \text { Mean } \circ (\mathcal{A}_{FAT} \cdot [  \boldsymbol{F}^{sem}_1\left(\Pi_1(\boldsymbol{x}), \boldsymbol{\theta}\right), \\
  & \cdots, \boldsymbol{F}^{sem}_N\left(\Pi_N(\boldsymbol{x}), \boldsymbol{\theta}\right)]^T)
 \end{aligned}
\vspace{-0.1in}
\end{equation}
The network detail of the Field-Aggregation Transformer can refer to the appendix.

\noindent \textbf{Joint Volumetric Rendering from both Fields.}  For radiance rendering, given a sequence of $\left\{\boldsymbol{f}^{rgb}_1, \cdots, \boldsymbol{f}^{rgb}_M\right\}$ from a sample ray, where $\boldsymbol{f}^{rgb}_i=\mathcal{F}^{rgb}\left(\boldsymbol{x}_i, \boldsymbol{\theta}\right) \in \mathbb{R}^{D_{rgb}}$  is the radiance feature of sampled points \(\boldsymbol{x}_i\) along its corresponding sample ray \(\boldsymbol{r}=(\boldsymbol{o}, \boldsymbol{d})\), we apply Ray-Aggregation Transformer (dubbed $\text{RAT}(\cdot)$) to aggregate weighted attention $\mathcal{A}_{RAT} \in \mathbb{R}^{N^{pts}}$ of the sequence to assemble the final feature vectors $S^{2D}_{rgb} \in \mathbb{R}^{D_{rgb}}$, then mean pooling and MLP layers are employed to map the feature vectors to RGB. The formulation of the above process is written below:
\begin{equation}
\begin{aligned}
    & S^{2D}_{rgb}(\boldsymbol{r}), \mathcal{A}_{RAT} = \text { RAT }(\boldsymbol{f}^{rgb}_1, \cdots, \boldsymbol{f}^{rgb}_M ) \\
    & C(\boldsymbol{r}) = \text{MLP} \circ \text{Mean} \circ S^{2D}_{rgb}(\boldsymbol{r})
\end{aligned}
\vspace{-0.1in}
\end{equation}
For semantic rendering, similar to the process of co-aggregate fields, given a sequence of  $\left\{\boldsymbol{f}^{sem}_1, \cdots, \boldsymbol{f}^{sem}_M\right\}$ from the same sampled ray,  we adopt dot-product between $\mathcal{A}_{RAT}$ and $\boldsymbol{f}^{sem}_i \in \mathbb{R}^{D_{sem}}$ to render semantic features $S^{2D}_{sem}(\boldsymbol{r}) \in \mathbb{R}^{D_{sem}}$ in novel view:
\begin{equation}
    S^{2D}_{sem}(\boldsymbol{r}) = \text{MLP} \circ \text { Mean } \circ (\mathcal{A}_{RAT} \cdot [\boldsymbol{f}^{sem}_1, \cdots, \boldsymbol{f}^{sem}_M]^T)
\end{equation}
The network detail of the Ray-Aggregation Transformer can be referred to the appendix.

\subsection{Optimizations}
\label{subsec:optimization}
We train the whole network from scratch under photometric loss $\mathcal{L}_{\text {rgb }}$, semantic pixel loss $\mathcal{L}_{\text {sem }}$ as well as our proposed semantic distill loss $\mathcal{L}^{2D}_{\text {disill}}$ and depth-guided semantic distill loss $\mathcal{L}^{dgs}_{\text{distill}}$, the overall loss $\mathcal{L}_{\text {all }}$ can be summarized as:
\begin{equation}
    \mathcal{L}_{\text {all }} = \alpha_1\cdot \mathcal{L}_{\text {rgb }} + \alpha_2 \cdot \mathcal{L}_{\text {sem }} + \alpha_3 \cdot \mathcal{L}^{2D}_{\text {distill}} + \alpha_4 \cdot \mathcal{L}^{dgs}_{\text {distill}}
\end{equation}

Photometric loss $\mathcal{L}_{\text {rgb }}$ and semantic pixel loss $\mathcal{L}_{\text {sem }}$ are pixel-level supervision, and they are widely used in NeRF and semantic tasks:
\begin{equation}
\mathcal{L}_{\text {rgb }}=\sum_{\mathbf{r} \in \mathcal{R}}\left\|\hat{\mathbf{C}}(\mathbf{r})-\mathbf{C}(\mathbf{r})\right\|_2^2,
\vspace{-0.1in}
\end{equation}
\begin{equation}
\mathcal{L}_{\text {sem }}=-\sum_{\mathbf{r} \in \mathcal{R}}\left[\sum_{l=1}^C p^c(\mathbf{r}) \log \hat{p}^c(\mathbf{r})\right],
\end{equation}
where $\mathcal{R}$ are the sampled rays within a training batch. $\hat{\mathbf{C}}(\mathbf{r}),\mathbf{C}(\mathbf{r})$ are the GT color and predicted color for ray $r$, respectively. Moreover, $p^c$ and $\hat{p^c}$  are the multi-class semantic probability at class \(c\) of the ground truth map.

\noindent \textbf{2D Semantic Distillation.} For semantic-driven tasks, it is crucial to augment the discrimination and semantic-aware ability of our rendered features.
Therefore, we propose \textbf{2D Semantic Distill Loss} $\mathcal{L}_{\text {S.D}}$. It distills~\cite{hinton2015distilling} the aggregated features \(\boldsymbol{S}^{2D}_{sem}\) by considering the features  \(\boldsymbol{S}^{2D}_{novel}\) extracted on novel-view as teacher, which effectively minimizes the differences between aggregated and teacher features:
\begin{equation}
\mathcal{L}_{\text {S.D}}=\sum_{\mathbf{r} \in \mathcal{R}} \left[1-\text{cos}\left(\boldsymbol{S}^{2D}_{sem}(r), \boldsymbol{S}^{2D}_{novel}(r)\right)\right]
\vspace{-0.1in}
\end{equation}

Since our model is trained from scratch, we apply a \textbf{gradient block} after ResNet-34 encoder to ensure that the loss function supervises the aggregation process of the Transformer modules to get better rendered semantic features \(\boldsymbol{S}^{2D}_{sem}\), otherwise, the extractor tends to learn less discriminative features to "cheat" the distillation loss.

\begin{figure}[t]
  \centering
  \includegraphics[width=1\linewidth]{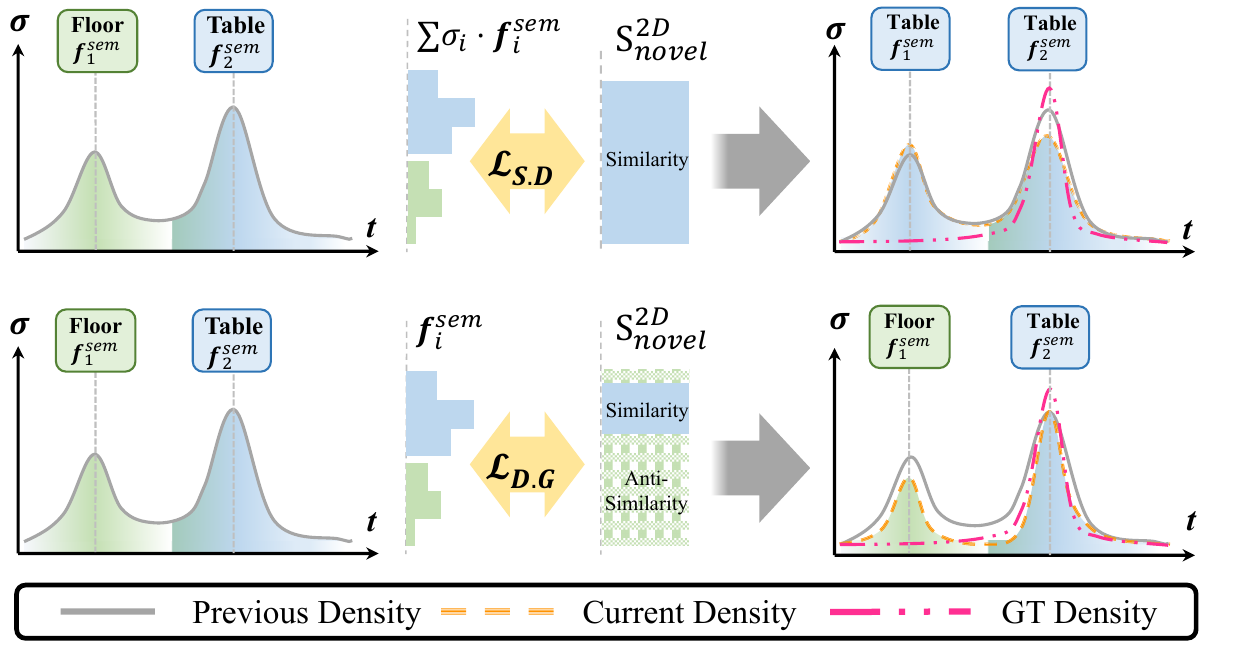}
 %  \vspace{-0.3in}
  \caption{\textbf{2D Semantic Distillation} $\mathcal{L}_{\text {S.D}}$ and \textbf{Depth-Guided Semantic Optimization} $\mathcal{L}_{\text {D.G}}$. This figure demonstrates a single raw of our semantic-embedding field.  the network "cheat" by rendering all points $\boldsymbol{f}^{sem}_i$ to the same prediction to satisfy $\mathcal{L}_{\text {S.D}}$ supervision. By performing spatial-wise semantic supervision,  $\mathcal{L}_{\text {S.D}}$ is able to mitigate the issue of "cheating".}
  \label{fig:dgs_loss}
  \vspace{-0.25in}
\end{figure}

\begin{table*}[t]
\centering
\small
\vspace{-0.1in}
\begin{tabular}{lll|c|ccl|ccl}
\toprule
\multicolumn{3}{l|}{}                           &                                  & \multicolumn{3}{c|}{Synthetic   Data (Replica~\cite{straub2019replica})}                    & \multicolumn{3}{c}{Real   Data (ScanNet~\cite{dai2017scannet})}                            \\ 
\multicolumn{3}{l|}{\multirow{-2}{*}{Method}}   & \multirow{-2}{*}{Settings}               & Total Acc$\uparrow$   & Avg Acc$\uparrow$  & mIoU$\uparrow$             & Total Acc$\uparrow$   & Avg Acc$\uparrow$   & mIoU$\uparrow$  \\ \midrule
% \multicolumn{3}{c|}{SemanticFPN}                &                                  & NaN          & NaN          & NaN          & 58.49          & 77.13        & 69.67        \\
\multicolumn{3}{l|}{MVSNeRF + Semantic Head}    &                                   & 54.25        & 33.70   & 23.41          & 60.01        & 46.01  & 39.82        \\
\multicolumn{3}{l|}{NeuRay + Semantic Head}     &                                   & 69.35        & 43.97   & 35.90          & 77.61        & 57.12  & 51.03        \\
\multicolumn{3}{l|}{Semantic-Ray}                      &                            & 70.51        & 47.19   & 41.59          & 78.24        & 62.55  & 57.15        \\
\rowcolor[HTML]{EFEFEF} 
\multicolumn{3}{l|}{\textbf{Ours}}              & \multirow{-5}{*}{Generalization}  & \textbf{78.01} & \textbf{50.80}  & \textbf{48.53}\textcolor{red}{$_{6.94\uparrow}$} &  \textbf{78.49} & \textbf{70.75} & \textbf{59.92}\textcolor{red}{$_{2.7\uparrow}$}\\ \midrule
% \multicolumn{3}{c|}{SemanticFPN}                &                                  & NaN          & NaN          & NaN          & NaN            & NaN          & NaN          \\
\multicolumn{3}{l|}{Semantic-NeRF}              &                                   & 94.36        & 70.20   & 75.06           & 97.54        & 93.89   & 91.24      \\
\multicolumn{3}{l|}{MVSNeRF + Semantic Headft}  &                                   & 79.48        & 62.85   & 53.77          & 76.25        & 69.70   & 55.26       \\
\multicolumn{3}{l|}{NeuRay +   Semantic Headft} &                                   & 85.54        & 70.05   & 63.73          & 91.56        & 81.04   & 77.48      \\
\multicolumn{3}{l|}{S-Rayft}                    &                                   & 96.38        & 80.81   & 75.96          & 98.20         & 93.97  & 91.06       \\
\rowcolor[HTML]{EFEFEF} 
\multicolumn{3}{l|}{\textbf{Oursft}}            & \multirow{-6}{*}{Finetuning}      & \textbf{97.60} & \textbf{86.45} & \textbf{87.72}\textcolor{red}{$_{11.76\uparrow}$}  & \textbf{98.43} & \textbf{94.77} & \textbf{93.84}\textcolor{red}{$_{2.78\uparrow}$}\\ \bottomrule
\end{tabular}
\vspace{-0.1in}
\caption{Quantitative Comparison with other SOTA methods for generalized and fine-tuning semantic segmentation. }
\vspace{-0.1in}
\label{tab:result}
\end{table*}

\begin{figure*}[t]
  \centering
    % \vspace{-0.2in}
    \centering\includegraphics[width=1\textwidth]{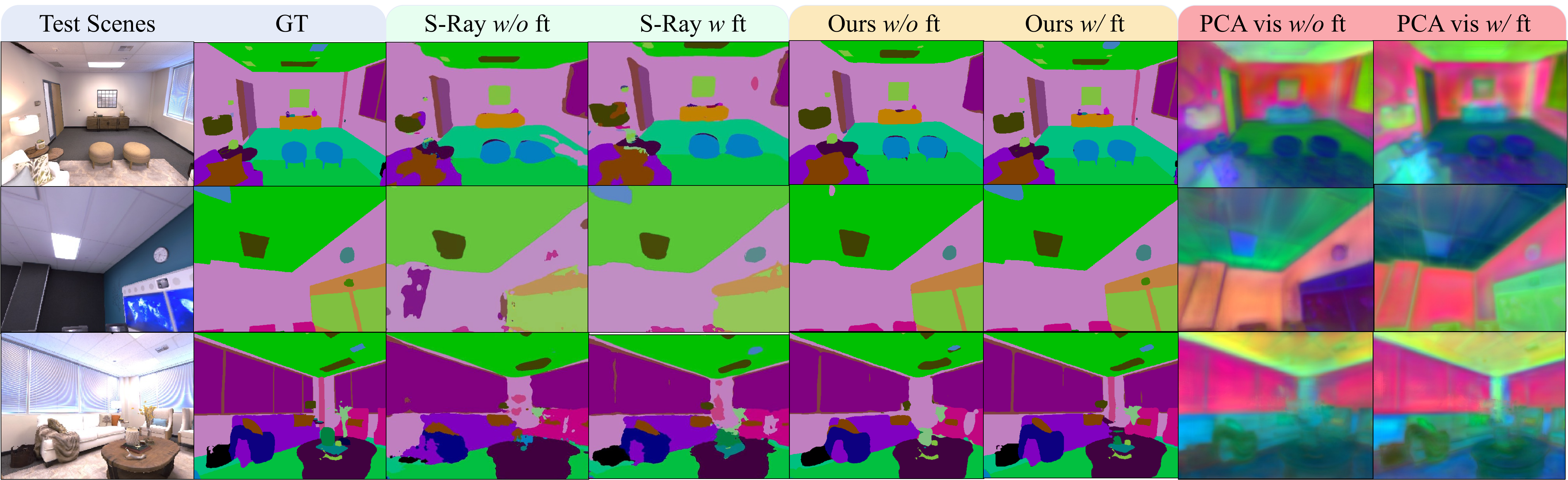}
      \vspace{-0.22in}
    \caption{\textbf{Semantic quality comparison in Replica}~\cite{straub2019replica}. On the left, we show the rendering results of S-Ray~\cite{liu2023semanticray} and GP-NeRF(ours) in generalized and finetuning settings. On the right, we visualize the PCA results of our rendered semantic features in novel views. }
  \label{fig:vis_semantic}
    \vspace{-0.21in}
\end{figure*}

\noindent \textbf{Depth-Guided Semantic Optimization.} 
It's worth noting that although $\mathcal{L}_{\text {S.D}}$ it significantly boosts the discrimination of rendered features, it also corrupts the geometry representation of our model. 
As illustrated in the first column of Fig. \ref{fig:dgs_loss}, the semantic representation of the ray is conducted by weighted summation of sampled point $\boldsymbol{f}^{sem}_i$ and their corresponding coefficient $\sigma_i$, where $\sigma_i$ belongs to $\mathcal{A}_{RAT}$ . Therefore, the loss can be minimized by misguiding $\boldsymbol{f}^{sem}_i$ (class 'Floor'\(\rightarrow\)'Table') rather than optimizing the attention weights $\sigma_i$(\textit{i.e. } geometry representation). To restore the semantic consistency with geometry constraint, we {proposed Depth-Guided Semantic Optimization} $\mathcal{L}_{\text{D.G}}$.  
given a sequence of sampled points $\boldsymbol{x}_i$  and corresponding features $\boldsymbol{f}^{sem}_i$ from ray $\mathbf{r}$, we perform per-point semantic distillation from the teacher's features ${S}^{2D}_{novel}(\boldsymbol{r})$:
\vspace{-0.1in}
\begin{equation}
    \mathcal{L}_{D.G}=\sum_{\mathbf{r} \in \mathcal{R}} \sum_{i=1}^{N_{pts}}{L_{sim}(x_i, \boldsymbol{f}^{sem}_{i},\boldsymbol{S}^{2D}_{novel}(\mathbf{r}))} 
\vspace{-0.1in}
\end{equation}
where $L_{sim}$ is the cosine embedding loss, it performs supervision under two situations:
(1) for those points $x_i$ near the GT depth ( $|x_i-x_d|<N_p$ ), it conducts similarity constraint with teacher features; (2) for those points far from the GT depth ($|x_i-x_d|>N_p$), it conducts anti-similarity constraint with teacher features, where $x_d$ is the sampled point projected by GT depth. In our implementation, $N_p$ is set to 2.
The formulation is shown below:

\vspace{-0.1in}
\begin{equation}
\small
L_{sim}(x_i, f_1, f_2)=\begin{cases}1-\cos \left(f_1, f_2\right)&,|x_i-x_d|<N_p\\ \max \left(0, \cos \left(f_1, f_2\right)\right)&,|x_i-x_d|>N_p\end{cases} 
\vspace{-0.1in}
\end{equation}
% where $f,n$ represent near depth and far depth.

\section{Experiments}

\subsection{Implementation Details}
We conduct experiments to compare our method against state-of-the-art methods for novel view synthesis with RGBs as well as semantic/instance labels. 
Firstly, we train our model in several scenes and directly evaluate our model on test scenes (i.e., unseen scenes).
Secondly, we finetune our generalized model on each unseen scene with small steps and compared them with per-scene optimized NeRF methods in semantic and reconstruction metrics. 
%Furthermore, we have conducted evaluations of our model on the task of segmenting individual 3D objects, which is aimed at demonstrating the model's capability to track unique objects within specific scenes.
%All results show our method achieves remarkable improvement compared with SOTA methods.

\label{sec:exp}
\noindent \textbf{Parameter Settings.}
We train our method end-to-end on datasets of multi-view posed images using the Adam optimizer to minimize the overall loss  $\mathcal{L}_{\text {all }}$. The learning rate or Multi-Task Feature Extractor, Transformer modules, and Perception Head are $5\times 10^{-3}$,$1\times 10^{-5}$ and $5\times 10^{-5}$ respectively, which decay exponentially over training steps. For generalized training, we train for 200,000 steps with 512 rays sampled in each iteration. For finetuning, we train for 10,000 steps for each scene.
Meanwhile, we sample 64 points per ray across all experiments. For each render interaction, we select \(N=10\) images as reference views. 

\noindent \textbf{Metrics.} Same as Semantic-Ray~\cite{liu2023semanticray}: (1) For semantic quality evaluation, we adopt mean Intersection-over-Union (mIoU) as well as average accuracy and total accuracy to compute segmentation quality. (2) For render quality evaluation, Peak Signal-to-Noise Ratio (PSNR), Structural Similarity Index Measure (SSIM)~\cite{wang2004image}, and the Learned Perceptual Image Patch Similarity (LPIPS)~\cite{zhang2018unreasonable} are adopted. More specifically, we refer to DM-NeRF~\cite{wang2022dm} and use AP of all 2D test images to evaluate instance quality evaluation.

\noindent \textbf{Datasets.} We train and evaluate our method on Replica~\cite{straub2019replica} and ScanNet~\cite{dai2017scannet} datasets. In these experiments, we use the same resolution and train/test splits as S-Ray~\cite{liu2023semanticray}.

\subsection{Comparison with State-of-the-Art}

\noindent \textbf{Generalized Semantic Results.} We compare our model with Semantic Ray, Generalized NeRFs(\textit{i.e.} NeuRay, MVSNeRF) with Semantic Head, and classical semantic segmentor (SemanticFPN) in both synthesis~\cite{straub2019replica} and real-world~\cite{dai2017scannet} datasets. We render the novel images in the resolution of \(640\times480\) for Replica, and \(320\times240\) for ScanNet.
As shown in Tab. \ref{tab:result}, our method achieves remarkable performance improvements compared with baselines. 
For example, our method significantly improves over Semantic-Ray by 6.94\% in Replica and 2.7\% in ScanNet. It's notable that Replica has more categories than ScanNet, and we achieve higher performance improvements in Replica, which further demonstrates the robustness and effectiveness of our semantic embedding field in handling complex semantic contexts.

\noindent \textbf{Fine-tuning Semantic Results.} We fine-tune our pre-trained with \textit{10k} steps for per-scene optimize evaluation. In Tab.~\ref{tab:result}, we observe that our method is superior to not only generalized methods but also per-scene optimization methods. Especially in ScanNet evaluation, we outperform the per-scene optimized method Semantic-NeRF~\cite{zhi2021semanticnerf} by a notable margin of 2.6\% in the mIoU metric. Comparatively, Semantic-Ray~\cite{liu2023semanticray} performs 0.18\% less effectively in the same metric.
Furthermore, the visual results in Fig. \ref{fig:vis_semantic} clearly reﬂect the quantitative results of Tab.~\ref{tab:result}. Given the benefit of jointly optimized attention maps to construct semantic embedding fields, our method demonstrates a clear ability to segment the boundaries of different classes effectively. This capability is particularly evident in the areas encircled in the figures.

\begin{table}[t]
\small
\begin{tabular}{c|cccl}
\toprule
Scene     & M-RCNN & Swin-T & DM-NeRF & Ours  \\
\midrule
Office\_0 & 74.05  & 80.17  & 82.71   & 90.46 \\
Office\_2 & 73.41  & 75.39  & 81.12   & 88.62 \\
Office\_3 & 72.91  & 73.26  & 76.30   & 82.64 \\
Office\_4 & 74.76  & 72.51  & 70.33   & 88.38 \\
Room\_0   & 78.67  & 76.90   & 79.83   & 89.91 \\
Room\_1   & 78.38  & 81.41  & 92.11   & 93.38 \\
Room\_2   & 77.58  & 80.33  & 84.78   & 93.11 \\
\rowcolor[HTML]{EFEFEF} 
Average   & 75.68  & 77.14  & 81.03   & 89.50\textcolor{red}{$_{8.47\uparrow}$} \\
\bottomrule
\end{tabular}
\vspace{-0.1in}
\caption{Quantitative results of instance segmentation results on Replica~\cite{straub2019replica}. The metric is AP$^{0.75}$.}
\vspace{-0.1in}
\label{tab:instance}
\end{table}

\begin{figure}[t]
  \centering
  \includegraphics[width=1\linewidth]{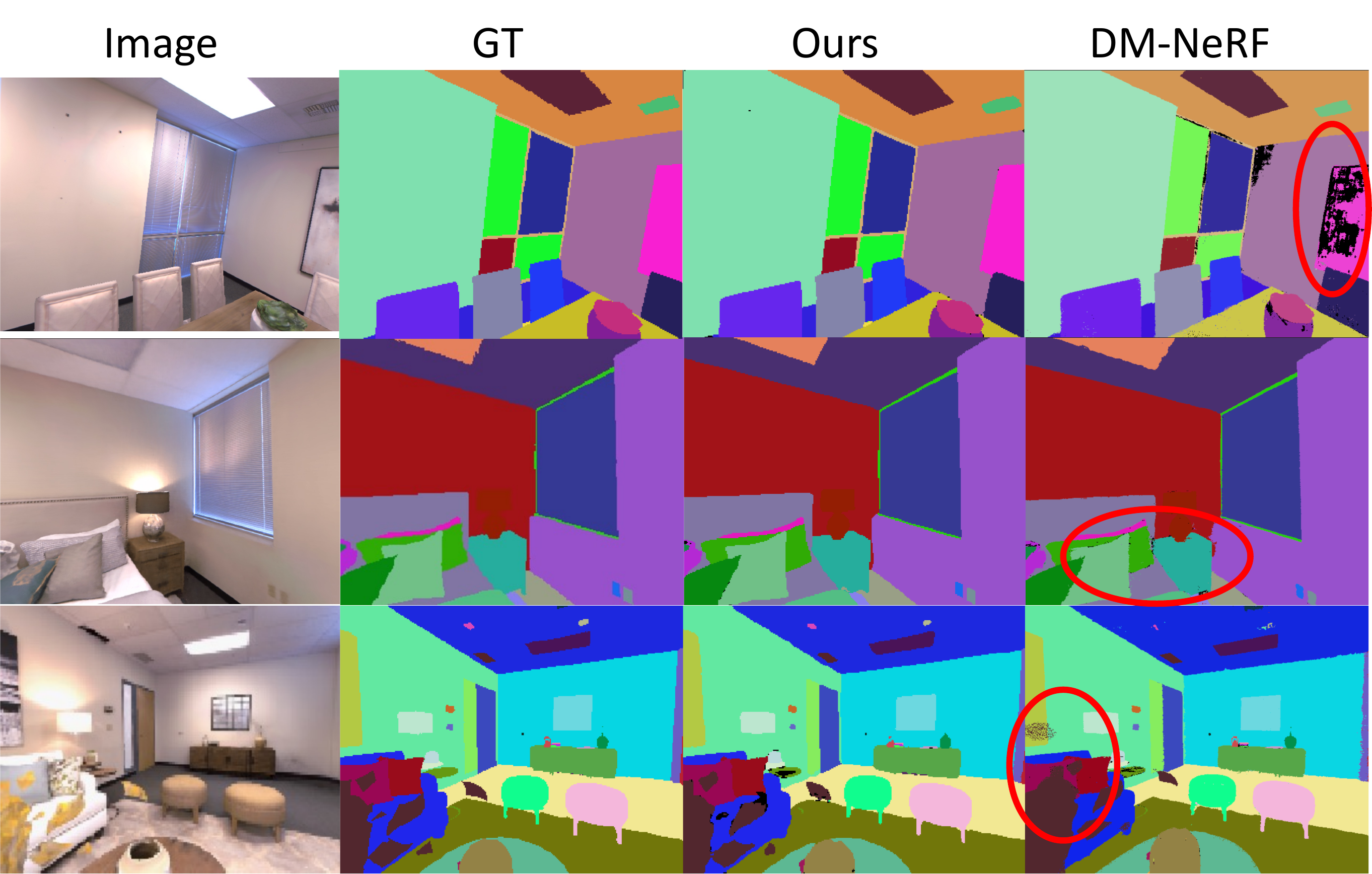}
   \vspace{-0.23in}
  \caption{Visualization of instance segmentation results on synthesis dataset~\cite{straub2019replica}. The discriminate area is highlighted with '{\color[HTML]{CB0000}$\bigcirc$}'.}
  \label{fig:visualization}
   \vspace{-0.1in}
\end{figure}

\noindent \textbf{Instance Segmentation Results.} With the success of our method in semantic scene representation, we explore the potential of our method in instance-level decomposition. 
Given the reason that the objects of each scene are unique, we only evaluate our performance in the per-scene optimization setting. 
Tab. \ref{tab:instance} presents the quantitative results. Not surprisingly, our method achieves excellent results for novel view prediction (+8.47\% \textit{w.r.t.} DM-NeRF~\cite{wang2022dm}) thanks to our powerful semantic embedding field and context-aware ability in novel view prediction. Figures \ref{fig:visualization}(a) further demonstrate that our semantic field can provide more discriminate semantic pattern than per-scene optimization method to decompose instances with accurate boundaries. Moreover, our method prevents the mis-segmentation of pixels within an instance thanks to our context-aware ability. These features enhance the accuracy and reliability of our scene perception process.

\begin{figure}[t]
  \centering
  \includegraphics[width=1\linewidth]{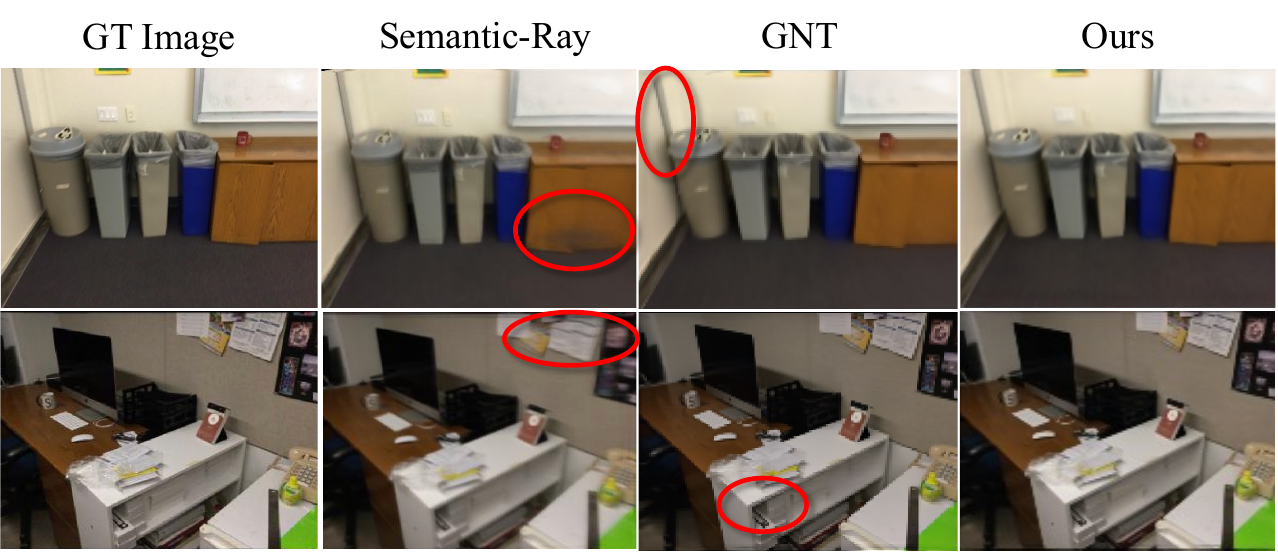}
    \vspace{-0.2in}
  \caption{Qualitative results of scene rendering for generalization settings in ScanNet~\cite{dai2017scannet}. We plot the discriminate area with '{\color[HTML]{CB0000}$\bigcirc$}'.}
    \vspace{-0.1in}
  \label{fig:psnr}
\end{figure}

\begin{table}[t]
\centering
\small
\begin{tabular}{l|lcc}
\toprule
Method        & PSNR$\uparrow$ & SSIM$\uparrow$ & LIPIPS$\downarrow$ \\ \midrule
Semantic-NeRF & 25.07 & 0.797 & 0.196   \\
MVSNeRF       & 23.84 & 0.733 & 0.267   \\
NeuRay        & 27.22 & 0.840  & 0.138   \\
Semantic-Ray         & 26.57 & 0.832 & 0.173   \\
Semantic-Ray$_{ft}$       & 29.27 & 0.865 & 0.127   \\
GNT           & 28.96 & 0.909   & 0.135     \\
GNT$_{ft}$        & 29.55 & 0.917   & 0.102     \\
\textbf{Ours}    & \textbf{29.37}\textcolor{red}{$_{2.8\uparrow}$} & \textbf{0.919} & \textbf{0.110} \\
\textbf{Ours$_{ft}$} & \textbf{29.60}\textcolor{red}{$_{0.33\uparrow}$}  & \textbf{0.923} & \textbf{0.102} \\ \bottomrule
\end{tabular}
\vspace{-0.1in}
\caption{Reconstruction Quality in ScanNet~\cite{dai2017scannet}. '\textit{ft}' denotes per-scene optimization using a generalized pre-trained model.}
\vspace{-0.25in}
\label{tab:psnr}
\end{table}

\noindent \textbf{Reconstruction Results.} It's worth noting that our method not only achieves SOTA in perception evaluation but also surpasses other SOTA methods in reconstruction quality. As shown in Tab \ref{tab:psnr}, in the generalized setting, our method surpasses Semantic-Ray~\cite{liu2023semanticray} by 2.8\% in PSNR, which is even better than Semantic-Ray with fine-tuning steps. 
Subsequently, we also improve the reconstruction quality by 0.41\% compared with GNT~\cite{t2023GNT} given the benefit on our radiance field is also supervised from semantic consistency. Fig. \ref{fig:psnr} provides visual evidence of our performance on ray rendering reconstruction, where our method delivers more detailed and clearer reconstruction results.

\subsection{Component Analysis and Ablation Study}
%We perform ablation analysis to understand the impact of each component of our method.

\noindent \textbf{Jointly Optimized Attention Maps.} 
As illustrated in sec. \ref{subsec:fields}, we aggregate semantic-embedding fields and render semantic features in novel views by sharing attention maps from Transformer modules.
In Tab. \ref{tab:ablation}, we compare the influence of our jointly optimized Field in ID. 1, 2, and evaluate their scene perception and reconstruction performances. In experiment ID. 1, when constructing the semantic field and aggregating features in novel views, we freeze the Attention maps from Transformers. Conversely, in experiment ID. 2, we unfreeze the attention maps and jointly optimize them through semantic and radiance supervision. 
Obviously, joint optimization can achieve better performance in semantic perception and ray reconstruction by 0.74\% and 0.24\%, compared with the frozen patterns. This approach further demonstrates that semantic consistency can provide a radiance reference for pixels within the same classes. Additionally, radiance consistency also contributes to achieving more accurate boundary segmentation. 

\begin{table}[t]
%\vspace{-0.1in}
\footnotesize
\begin{tabular}{ccccccc}
\toprule
\multirow{2}{*}{ID} &
  \multirow{2}{*}{\begin{tabular}[c]{@{}c@{}}Jointly \\ Optimized\end{tabular}} &
  \multirow{2}{*}{\begin{tabular}[c]{@{}c@{}}2D S.D \\ Loss\end{tabular}} &
  \multirow{2}{*}{\begin{tabular}[c]{@{}c@{}}Gradient\\ Block\end{tabular}} &
  \multirow{2}{*}{\begin{tabular}[c]{@{}c@{}}D.G \\ Loss\end{tabular}} &
  \multirow{2}{*}{mIoU↑} &
  \multirow{2}{*}{PSNR↑} \\
  &   &   &   &   &       &       \\
\midrule
1 & {\color[HTML]{CB0000}\XSolidBrush} & {\color[HTML]{CB0000}\XSolidBrush} & {\color[HTML]{CB0000}\XSolidBrush} & {\color[HTML]{CB0000}\XSolidBrush} & 56.45 & 29.06  \\
2 & \color[HTML]{009901}\textbf{\Checkmark} & {\color[HTML]{CB0000}\XSolidBrush} & {\color[HTML]{CB0000}\XSolidBrush} & {\color[HTML]{CB0000}\XSolidBrush} & 57.19 & 29.30 \\
3 & \color[HTML]{009901}\textbf{\Checkmark} & \color[HTML]{009901}\textbf{\Checkmark} & {\color[HTML]{CB0000}\XSolidBrush} & {\color[HTML]{CB0000}\XSolidBrush} & 52.03 & 29.29 \\
4 & \color[HTML]{009901}\textbf{\Checkmark} & \color[HTML]{009901}\textbf{\Checkmark} & \color[HTML]{009901}\textbf{\Checkmark} & {\color[HTML]{CB0000}\XSolidBrush} & 59.55 & 29.26 \\
5 & \color[HTML]{009901}\textbf{\Checkmark} & \color[HTML]{009901}\textbf{\Checkmark} & \color[HTML]{009901}\textbf{\Checkmark} & \color[HTML]{009901}\textbf{\Checkmark} & 59.92 & 29.37 \\
\bottomrule
\end{tabular}
 \vspace{-0.1in}
\caption{Ablations of our design choices on ScanNet~\cite{dai2017scannet}. Notice that 'Gradient Block' is dependent on '2D S.D Loss' and 'D.G Loss', where 2D S.D denotes 2D Semantic Distill Loss and D.G denotes Depth-Guided Semantic Enhancement.}
\label{tab:ablation}
\vspace{-0.1in}
\end{table}

\begin{figure}[t]
  \centering
  \begin{subfigure}{0.49\linewidth}
    \includegraphics[width=1\linewidth]{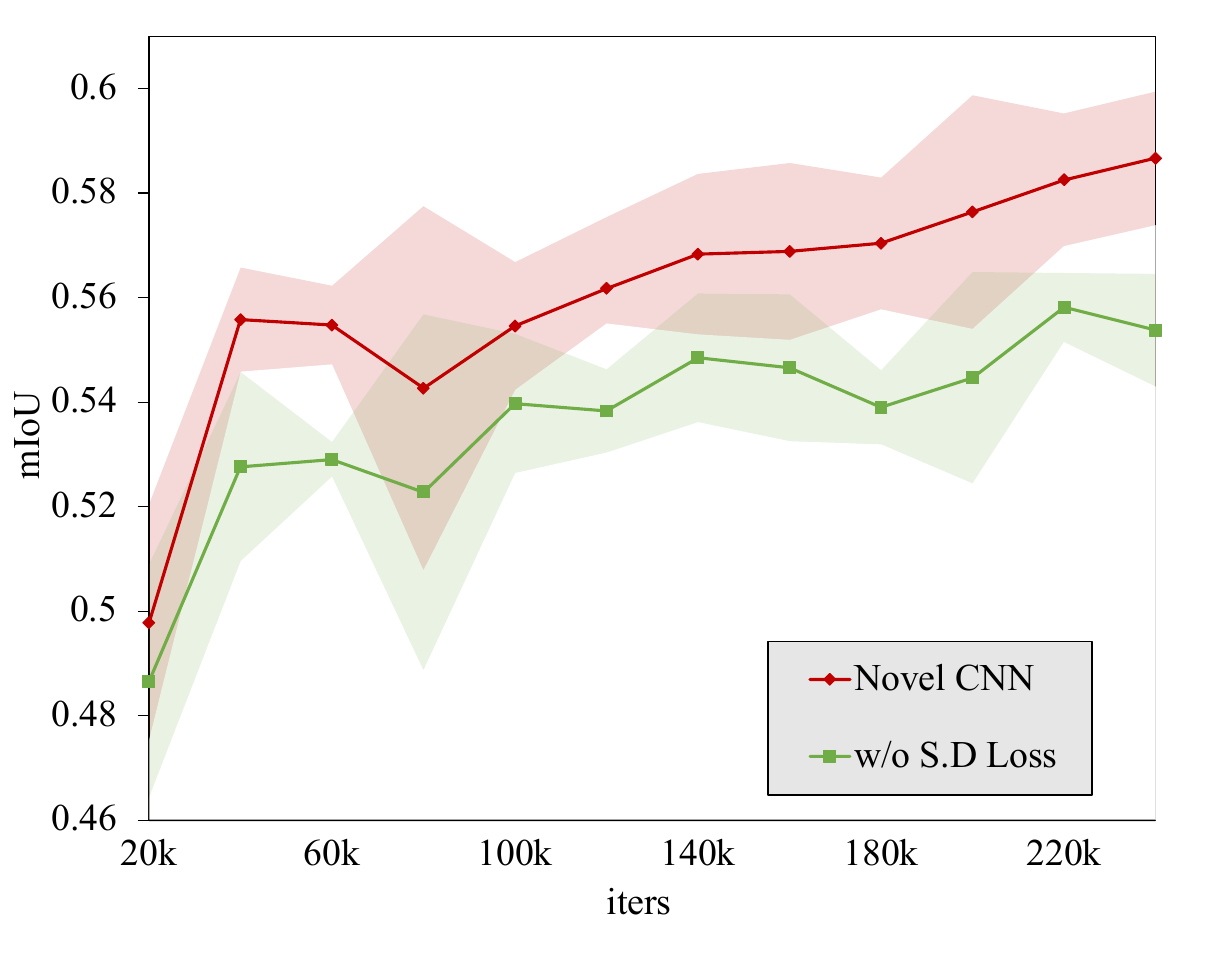}
    \caption{ \textbf{\textit{w/o}} S.D Loss and gradient block.}
    \label{fig:pseudo-our}
  \end{subfigure}
  \hfill
  \begin{subfigure}{0.49\linewidth}
    \includegraphics[width=1\linewidth]{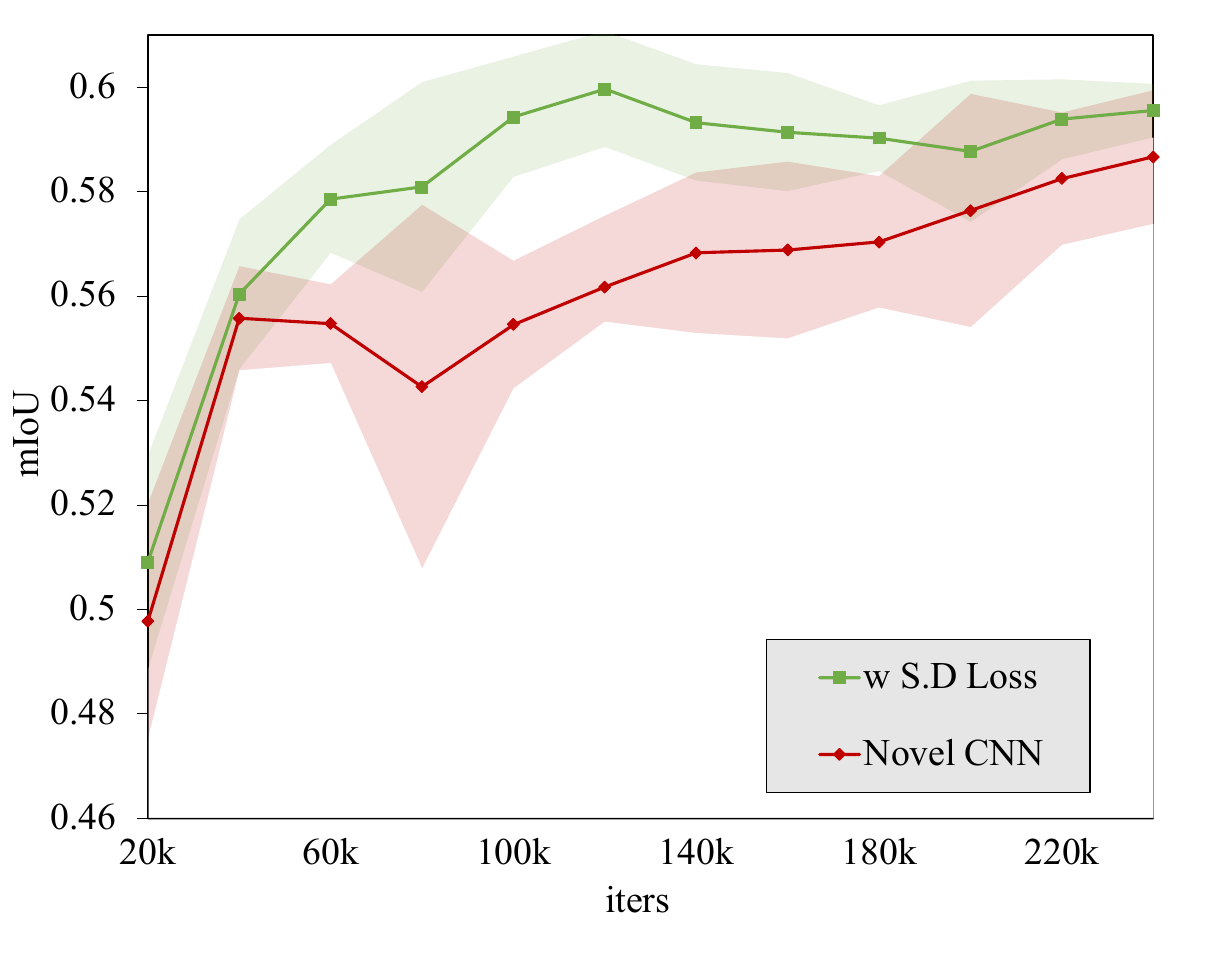}
    \caption{ \textbf{\textit{w/}} S.D Loss and gradient block.}
    \label{fig:pseudo-freesolo}
  \end{subfigure}
   \vspace{-0.1in}
  \caption{Ablations of \textbf{Semantic Distillation Loss via Gradient Block}. \textcolor{red}{Red} part denotes the mIoU results predicted by extracted features from novel image. \textcolor{green}{Green} part denotes the mIoU predicted by the rendered features from semantic embedding fields.}
  \label{fig:distill_ablation}
  \vspace{-0.2in}
\end{figure}

\noindent \textbf{Semantic Distill Loss and Gradient Block.} ID. 3, 4 in Tab. \ref{tab:ablation} reflect the influence of 2D semantic distillation loss and corresponding gradient block. As observed, there is a significant drop in performance (-5.16 compared to ID. 2) when only the 2D semantic distillation loss is adopted, which means the shared parts of the teacher and student branch (\textit{i.e.} CNN encoder and FPN) tend to learn less discriminate features to "cheat" the distillation loss. Meanwhile, with our \textbf{Gradient Block}, the situation can be solved, and the performance of mIoU achieves remarkable improvements by 7.52\%. 
Moreover, we repeat the ID.3, 4 experiments five times and show the mIoU learning curves on ScanNet~\cite{dai2017scannet} in Fig.~\ref{fig:distill_ablation}. We can observe that this contribution leads to a more precise convergence speed and higher final accuracy (See Fig. \ref{fig:distill_ablation}(b)).

\noindent \textbf{Depth-Guided Semantic Distill Loss.} 
It is notable that 2D semantic distill has a negative impact on reconstruction quality, by 0.4\% in PSNR compared with ID. 2, which is due to the fact that the 2D semantic distill loss can only supervise the rendered features rather than 3D points within the rays. Under this circumstance, some points in the ray would be "cheated" by adjusting the semantic representation to satisfy distillation loss, which would further impact the actual weight distribution of the points in sample rays. 
ID. 5 in Fig. \ref{tab:ablation} shows that $\mathcal{L}_{D.G}$ yields clear improvement by 0.37\% and 0.11\% in mIoU and PSNR, indicating that a more precise, 3D-level semantic supervision can partially improve the geometry awareness of our semantic field and suppress the "cheating" phenomenon.

\section{Conclusion}
In this paper, we propose GP-NeRF, the first unified learning framework that combines NeRF and segmentation modules to perform context-aware 3D scene perception. Unlike previous NeRF-based approaches that render semantic labels for each pixel individually, the proposed GP-NeRF utilizes many contextual modeling units from the widely-studied 2D segmentors and introduces Transformers to co-construct radiance as well as semantic embedding fields and facilitates the joint volumetric rendering upon both fields for novel views. New self-distillation mechanisms are also designed to further boost the quality of the semantic embedding field. Comprehensive experiments demonstrate that GP-NeRF achieves significant performance improvements (sometimes $>10\%$) compared to existing SOTA methods. In the future, we will follow more recent studies on visual saliency~\cite{liu2020picanet,suo2023text,guo2023semantic} as well as semi-supervised learning techniques \cite{chen2024virtual, liu2022learning,xue2023weakly,zhang2023weakly} to overcome the scenario of lack of full annotated semantic labels~\cite{li2023boosting,cheng2023hybrid,zhang2022generalized,liang2024caphuman} in scene understanding task.

\section{Acknowledgement}

This work is supported in part by the National Natural Science Foundation of China (No. 62293543, 62322605, U21B2048), and by the Institute of Artificial Intelligence,Hefei Comprehensive National  Science Center Project under Grant (21KT008).

%%%%%%%%% REFERENCES
{\small
\bibliographystyle{ieee_fullname}
\bibliography{PaperForReview}
}

\clearpage  

\maketitlesupplementary

\subsection{Implementation of our Transformers}

We provide a simple and efﬁcient pytorch pseudo-code to implement the attention operations in the field-aggregation, ray-aggregation transformer blocks in Alg. 1, 2.
We use ray features to generate attention maps $A_{field}$ and $A_{ray}$, and reuse them to construct semantic-embedding field as well as semantic features rendering.

\begin{algorithm}
\label{alg:1}
\small
\caption{Field-Aggregation Transformer} 
\SetKwInput{KwNetwork}{Network}
\SetKwInput{KwForward}{Forward}

\KwIn{
\\$\boldsymbol{X}_0 \rightarrow$ coordinate aligned features $\left(N_{\text {rays }}, N_{\mathrm{pts}}, D_{rgb}\right)$ \\
\begin{minipage}{1.0\linewidth}
$\boldsymbol{X}_{rgb} \rightarrow$ epipolar view Ray feats $\left(N_{\text {rays }}, N_{\mathrm{pts}}, N_{\mathrm{views}}, D_{rgb}\right)$ \\ 
$\boldsymbol{X}_{sem} \rightarrow$ epipolar view Sem feats $\left(N_{\text {rays }}, N_{\mathrm{pts}}, N_{\mathrm{views}}, D_{sem}\right)$ \\ 
\end{minipage} 
$\boldsymbol{\Delta} \boldsymbol{d} \rightarrow$ relative directions $\left(N_{\mathrm{rays}}, N_{\mathrm{pts}}, N_{\text {views }}, 3\right)$ \\
}
\KwNetwork{$f_Q, f_K, f_V, f_P, f_A, f_{rgb} \rightarrow$ MLP layers}
\KwOut{\(\boldsymbol{S}^{3D}_{rgb}, \boldsymbol{S}^{3D}_{sem}\)} 
\vspace{1mm} \hrule \vspace{1mm}
\KwForward{\sethlcolor{lightgray}\hl{{\color[RGB]{232, 0, 0}Red}} for semantic-embedding field aggregation}
\vspace{1mm} \hrule \vspace{1mm}
$\boldsymbol{Q}=f_Q\left(\boldsymbol{X}_0\right), \boldsymbol{K}=f_K\left(\boldsymbol{X}_{rgb}\right), \boldsymbol{V}=f_V\left(\boldsymbol{X}_{rgb}\right)$ \\
$\boldsymbol{P}_{field}=f_P(\boldsymbol{\Delta} \boldsymbol{d})$ \\
$\boldsymbol{A}_{field}=\boldsymbol{K}-\boldsymbol{Q}[:,:$ None, $:]+\boldsymbol{P}$ \\
$\boldsymbol{A}_{field}=\operatorname{softmax}(\boldsymbol{A}, \operatorname{dim}=-2)$ \\
\sethlcolor{lightgray}\hl{${\color[RGB]{232, 0, 0}\boldsymbol{A}'_{field} = \boldsymbol{A}_{field}\cdot\text{repeat\_interleave}(4)}$}\\
\sethlcolor{lightgray}\hl{${\color[RGB]{232, 0, 0}\boldsymbol{P}'_{field} = \boldsymbol{P}\cdot\text{repeat\_interleave}(4)}$}\\
$\boldsymbol{S}^{3D}_{rgb}=((\boldsymbol{V}+\boldsymbol{P}) \cdot \boldsymbol{A}) \cdot \operatorname{sum}(\operatorname{dim}=2)$\\
$\boldsymbol{S}^{3D}_{rgb}=f_{rgb}(\boldsymbol{\boldsymbol{S}^{3D}_{rgb}})$\\
\sethlcolor{lightgray}\hl{${\color[RGB]{232, 0, 0}\boldsymbol{S}^{3D}_{sem}=((\boldsymbol{X}_{sem}+\boldsymbol{P}'_{field})  \cdot \boldsymbol{A}'_{field})\cdot \operatorname{sum}(\operatorname{dim}=2)}$}
\end{algorithm}

\begin{algorithm}
\label{alg:2}

\small
\caption{Ray-Aggregation Transformer} 
\SetKwInput{KwNetwork}{Network}
\SetKwInput{KwForward}{Forward}

\KwIn{\\
\begin{minipage}{1.0\linewidth}
$\boldsymbol{X}^{rgb}_0 \rightarrow$ coordinate aligned rgb features $\left(N_{\text {rays }}, N_{\text {pts }}, D_{rgb}\right)$\\
$\boldsymbol{X}^{sem}_0 \rightarrow$ coordinate aligned sem features $\left(N_{\text {rays }}, N_{\text {pts }}, D_{sem}\right)$\\
$\boldsymbol{x} \rightarrow$ point coordinates (after PE) $\left(N_{\text {rays }}, N_{\text {pts }}, D_{rgb}\right)$\\
$d \rightarrow$ target view direction (after PE) $\left(N_{\text {rays }}, N_{\text {pts }}, D_{rgb}\right)$
\end{minipage}
}
\KwNetwork{$f_Q, f_K, f_V, f_P, f_A, f_{rgb}, f_{sem} \rightarrow$ MLP layers}
\KwOut{\(\boldsymbol{S}^{2D}_{rgb}, \boldsymbol{S}^{2D}_{sem}\)} 
\vspace{1mm} \hrule \vspace{1mm}
\KwForward{\sethlcolor{lightgray}\hl{{\color[RGB]{232, 0, 0}Red}} for semantic-embedding field aggregation}
\vspace{1mm} \hrule \vspace{1mm}
$\boldsymbol{X}^{rgb}_0 = f_P(\text{concat}(\boldsymbol{X}^{rgb}_0,d,x))$\\
$\boldsymbol{Q}=f_Q\left(\boldsymbol{X}^{rgb}_0\right), \boldsymbol{K}=f_K\left(\boldsymbol{X}^{rgb}_0\right), \boldsymbol{V}=f_V\left(\boldsymbol{X}^{rgb}_0\right)$ \\

$\boldsymbol{A}_{ray}=\operatorname{matmul}\left(\boldsymbol{Q}, \boldsymbol{K}^T\right) / \sqrt{D}$ \\ 
$\boldsymbol{A}_{ray}=\operatorname{softmax}(\boldsymbol{A}_{ray}, \operatorname{dim}=-1)$ \\
\sethlcolor{lightgray}\hl{${\color[RGB]{232, 0, 0}\boldsymbol{A}'_{ray} = \boldsymbol{A}_{ray}\cdot\text{repeat\_interleave}(4)}$}\\

$\boldsymbol{S}^{2D}_{rgb}=\operatorname{matmul}(\boldsymbol{V}, \boldsymbol{A}_{ray})$ \\
$\boldsymbol{S}^{2D}_{rgb}=f_{rgb}(\boldsymbol{S}^{2D}_{rgb})$ \\

\sethlcolor{lightgray}\hl{${\color[RGB]{232, 0, 0}\boldsymbol{S}^{2D}_{sem}=\operatorname{matmul}(\boldsymbol{X}^{sem}_0, \boldsymbol{A}'_{ray})}$}
\end{algorithm}

\subsection{Reconstruction results in instance setting}
During the novel view instance segmentation task, we evaluate our reconstruction results and compare them with SOTA method DM-NeRF\cite{wang2022dm}. 
As shown in Table \ref{tab:instance-psnr}, our approach surpasses DM-NeRF in terms of SSIM and LPIPS metrics by 0.02\% and 0.065\%, respectively. It demonstrates that contextual information from semantic features can enhance the geometry reconstruction in our jointly optimized field and rendering framework.

\begin{table}[htbp]
\tiny
\caption{Quantitative results of reconstruction task in Replica\cite{straub2019replica} during instance segmentation setting.}\vspace{-8pt}
\begin{tabular}{c|ccc|ccc}
\toprule
                        & \multicolumn{3}{c|}{DM-NeRF} & \multicolumn{3}{c}{Ours} \\
\multirow{-2}{*}{Scene} & PSNR$\uparrow$    & SSIM$\uparrow$    & LPIPS$\downarrow$   & PSNR$\uparrow$   & SSIM$\uparrow$   & LPIPS$\downarrow$  \\
\midrule
Office\_0               & 40.66   & 0.972   & 0.07    & 39.25  & 0.984  & 0.027  \\
Office\_2               & 36.98   & 0.964   & 0.115   & 36.01  & 0.974  & 0.042  \\
Office\_3               & 35.34   & 0.955   & 0.078   & 36.02  & 0.982  & 0.027  \\
Office\_4               & 32.95   & 0.921   & 0.172   & 32.75  & 0.94   & 0.085  \\
Room\_0                 & 34.97   & 0.94    & 0.127   & 34.29  & 0.972  & 0.049  \\
Room\_1                 & 34.72   & 0.931   & 0.134   & 36.45  & 0.968  & 0.043  \\
Room\_2                 & 37.32   & 0.963   & 0.115   & 34.75  & 0.960  & 0.085  \\
\rowcolor[HTML]{EFEFEF} 
Average                 & 36.13   & 0.949   & 0.116   & 35.64\textcolor{green}{$_{0.49\downarrow}$}& 0.969\textcolor{red}{$_{0.02\uparrow}$}  & 0.051\textcolor{red}{$_{0.065\downarrow}$} \\
\bottomrule
\end{tabular}
\label{tab:instance-psnr}
\end{table}

\subsection{Few-step Finetuning Comparison} Tab. \ref{tab:time} presents a comparison of different models, showcasing their mIoU and finetuning times on the ScanNet~\cite{dai2017scannet} dataset, along with the AP75 metric in Replica~\cite{straub2019replica}. We observe that by ﬁnetuning with limited time, our model is able to achieve a better perception accuracy than a well-trained per-scene optimized method, such as 3.45\% in mIoU with Semantic-NeRF~\cite{zhi2021semanticnerf} and 3.7\% in AP75 with DM-NeRF~\cite{wang2022dm}. Specifically, we observe that our method surpasses Semantic-Ray, requiring only half as many finetuning steps, and improves the mIoU by 0.74\%, which further demonstrates that our semantic embedding field with more discrimination successfully improves the generalized ability.

\begin{table}[htbp]
\small
\begin{tabular}{cccc}
\toprule
Method         & Train Step & Train Time  & mIoU/AP75  \\
\midrule
Semantic-NeRF~\cite{zhi2021semanticnerf}  & 50k        & $\sim$2h    & 89.33 \\
MVSNeRF w/s-Ft & 5k         & $\sim$20min & 52.02 \\
NeuRay~\cite{liu2022neuray} w/s-Ft  & 5k         & $\sim$32min & 79.23 \\
Semantic-Ray~\cite{liu2023semanticray}-Ft       & 5k         & $\sim$20min & 92.04 \\ 
Ours-Ft        & 2.5k       & $\sim$20min & 92.78\textcolor{red}{$_{0.74\uparrow}$} \\ \midrule
 DM-NeRF     & 200k       & $\sim$2h    & 81.03     \\
\rowcolor[HTML]{EFEFEF} 
 Ours-Ft     & 4k         & $\sim$30min & 84.73\textcolor{red}{$_{3.7\uparrow}$}  \\
\bottomrule
\end{tabular}
\caption{mIoU and training steps/time on ScanNet~\cite{dai2017scannet}. "w/ s" means adding a semantic head on the baseline architectures.}\vspace{-8pt}
\label{tab:time}
\end{table}

We further evaluate the above experiments in instance segmentation setting, shown in the bottom column in Tab. \ref{tab:time}. Not surprising, compared with SOTA method DM-NeRF\cite{wang2022dm}, we achieve better performance with only 4k training steps, by 3.7\% in AP75. 
% \begin{table}[t]
% \small
% \begin{tabular}{c|cccc}
% \toprule
% \multirow{2}{*}{Model iters} & \multicolumn{4}{c}{AP75}                  \\
%                              & Office\_2 & Office\_4 & Room\_0 & Room\_1 \\
% \midrule
% DM-NeRF 200k                 & 81.12     & 70.33     & 79.83   & 92.11   \\
% Ours 2k                      & 77.53     & 66.45     & 74.05   & 90.12   \\
% Ours 4k                      & 83.91     & 79.31     & 93.11   & 93.38   \\
% Ours 6k                      & 82.2      & 82.64     & 93.06   & 95.37   \\
% Ours 8k                      & 85.89     & 86.27     & 93.61   & 94.87   \\
% Ours 10k                     & 88.38     & 86.37     & 93.59   & 95.50   \\
% Ours 12k                     & 89.9      & 88.87     & 93.49   & 94.91   \\
% \bottomrule
% \end{tabular}
% \end{table}

\begin{figure*}[t]
    \includegraphics[width=1\linewidth]{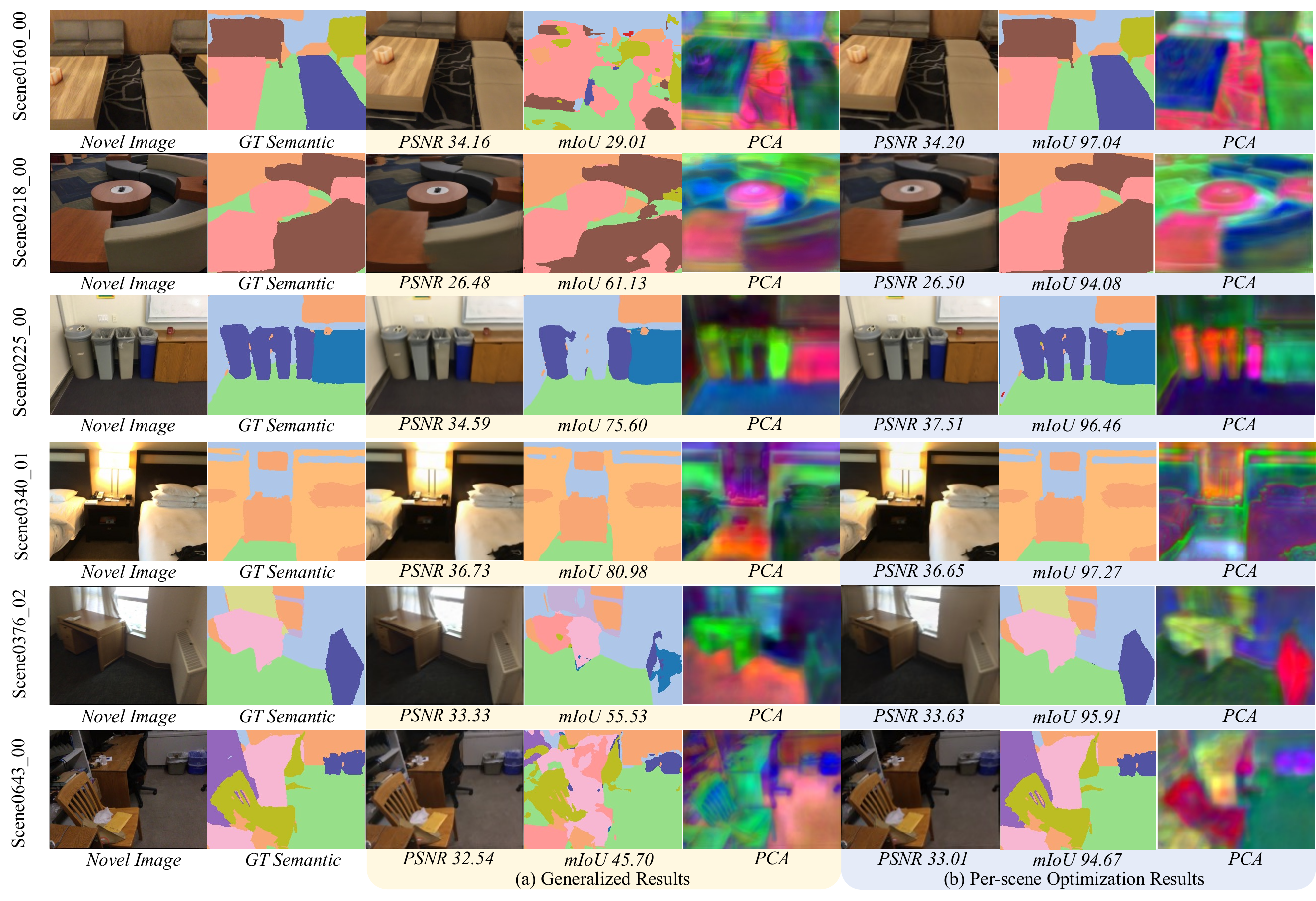}
    \caption{The visualization results in ScanNet\cite{dai2017scannet}. Here we visualize the semantic as well as reconstruction results in both generalized and finetuning settings. }
    \label{fig:supp}
\end{figure*}
\subsection{Additional Visualization Results}
Fig. \ref{fig:supp} shows the additional qualitative results of semantic prediction and reconstruction.

\end{document}